\documentclass[journal]{IEEEtran}

\usepackage{fancyhdr}
\usepackage{cite}
\usepackage{amsmath,amssymb,bm}
\usepackage{array}
\ifCLASSOPTIONcompsoc
  \usepackage[caption=false,font=normalsize,labelfont=sf,textfont=sf]{subfig}
\else
  \usepackage[caption=false,font=footnotesize]{subfig}
\fi
\usepackage{url}
\usepackage{algorithm}
\usepackage{balance}
\usepackage{color}
\usepackage{epsfig}
\usepackage{epstopdf}
\usepackage{enumitem}
\usepackage{float}
\usepackage{graphicx}
\usepackage{setspace}
\usepackage{wasysym}
\usepackage{galois}
\setenumerate[1]{itemsep=0pt,partopsep=0pt,parsep=\parskip,topsep=0pt}

\newfloat{figtab}{htb}{f}
\makeatletter
  \newcommand\figcaption{\def\@captype{figure}\caption}
  \newcommand\tabcaption{\def\@captype{table}\caption}
\makeatother
\usepackage[colorlinks, linkcolor=black, anchorcolor=black, citecolor=black]{hyperref}

\begin{document}
	
\title{Ratio-Preserving Half-Cylindrical Warps for Natural Image Stitching}
\author{Yifang Xu, \thanks{Y. Xu is with the Center for Combinatorics, Nankai University, Tianjin 300071, China. Email: xyf@mail.nankai.edu.cn.}
	 Jing Chen, \thanks{J. Chen is with the Center for Combinatorics, Nankai University, Tianjin 300071, China. Email: chenjing@mail.nankai.edu.cn.}  
	 Tianli Liao$^*$ \thanks{T. Liao is with the Center for Combinatorics, Nankai University, Tianjin 300071, China. Email: liaotianli@mail.nankai.edu.cn.}}

%\thanks{T. Liao is with the Center for Combinatorics, Nankai University, Tianjin 300071, China. Email: liaotianli@mail.nankai.edu.cn.}

\maketitle

\begin{abstract}
A novel warp for natural image stitching is proposed that utilizes the property of cylindrical warp and a horizontal pixel selection strategy. The proposed ratio-preserving half-cylindrical warp is a combination of homography and cylindrical warps which guarantees alignment by homography and possesses less projective distortion by cylindrical warp. Unlike previous approaches applying cylindrical warp before homography, we use partition lines to divide the image into different parts and apply homography in the overlapping region while a composition of homography and cylindrical warps in the non-overlapping region. The pixel selection strategy then samples the points in horizontal and reconstructs the image via interpolation to further reduce horizontal distortion by maintaining the ratio as similarity. With applying half-cylindrical warp and horizontal pixel selection, the projective distortion in vertical and horizontal is mitigated simultaneously. Experiments show that our warp is efficient and produces a more natural-looking stitched result than previous methods.
\end{abstract}


\IEEEpeerreviewmaketitle

\section{Introduction}
\begin{figure}[h]
	\centering{\includegraphics[width=0.485\textwidth]{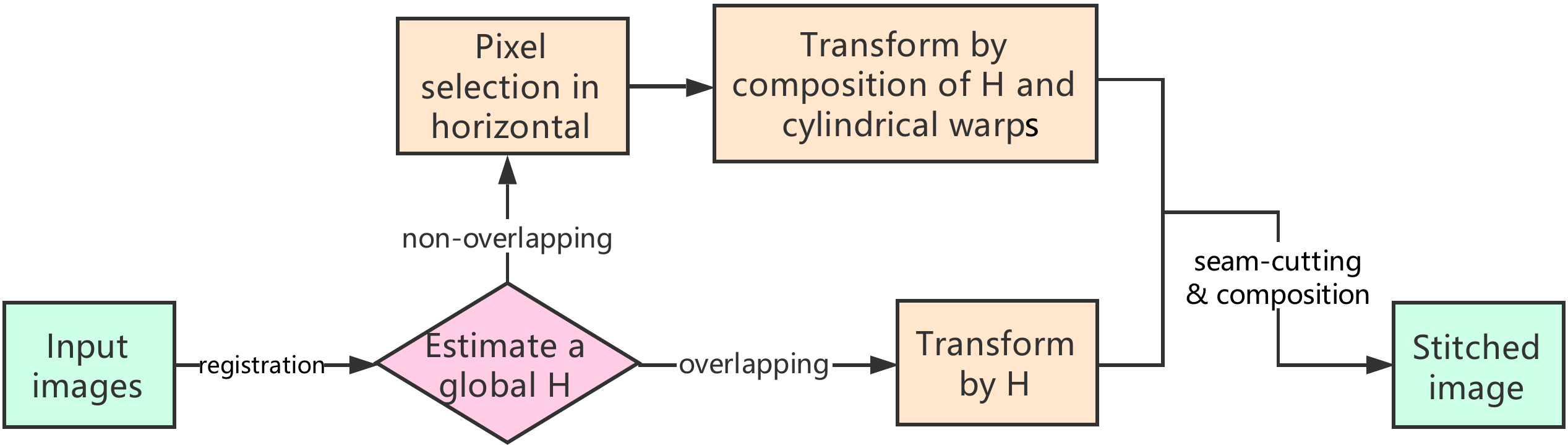}}
	\caption{Overview of ratio-preserving half-cylindrical warp for image stitching}
	\label{overview}
\end{figure}

Image stitching has been well studied and widely applied in computer vision, where the most crucial step is to determine a 2D warp for each image and transform it into a common plane. Among these warps, similarity warp can preserve the angles between lines and maintain a uniform scale factor (ratio).  Homography is the most flexible and popular global warp which preserves all straight lines, however, it introduces projective distortion at the same time. Basically, the projective distortion usually consists of distortion in vertical and horizontal. Many efforts have been devoted to mitigating these projective distortion. AutoStitch \cite{1} uses cylindrical warp as pre-processing step to address this problem, however, it inevitably curves straight lines and suffers from horizontal distortion. SPHP \cite{2} takes advantage of similarity to maintain the original perspective of input image but suffers from line-bending. QH \cite{3} uses the proposed quasi-homography warp to mitigate projective and perspective distortion simultaneously, but there still exists horizontal stretches and vertical distortion. In this letter, an effective algorithm based on the property of cylindrical warp is proposed to mitigate the projective distortion produced by homography. The strategy of image resizing for seam-cutting significantly decreases the processing time in different resolutions. Fig. \ref{overview} shows the flowchart of our proposed algorithm.

\section{Proposed half-cylindrical warp}

\begin{figure}[h]
	\centering
	\includegraphics[width=0.46\textwidth]{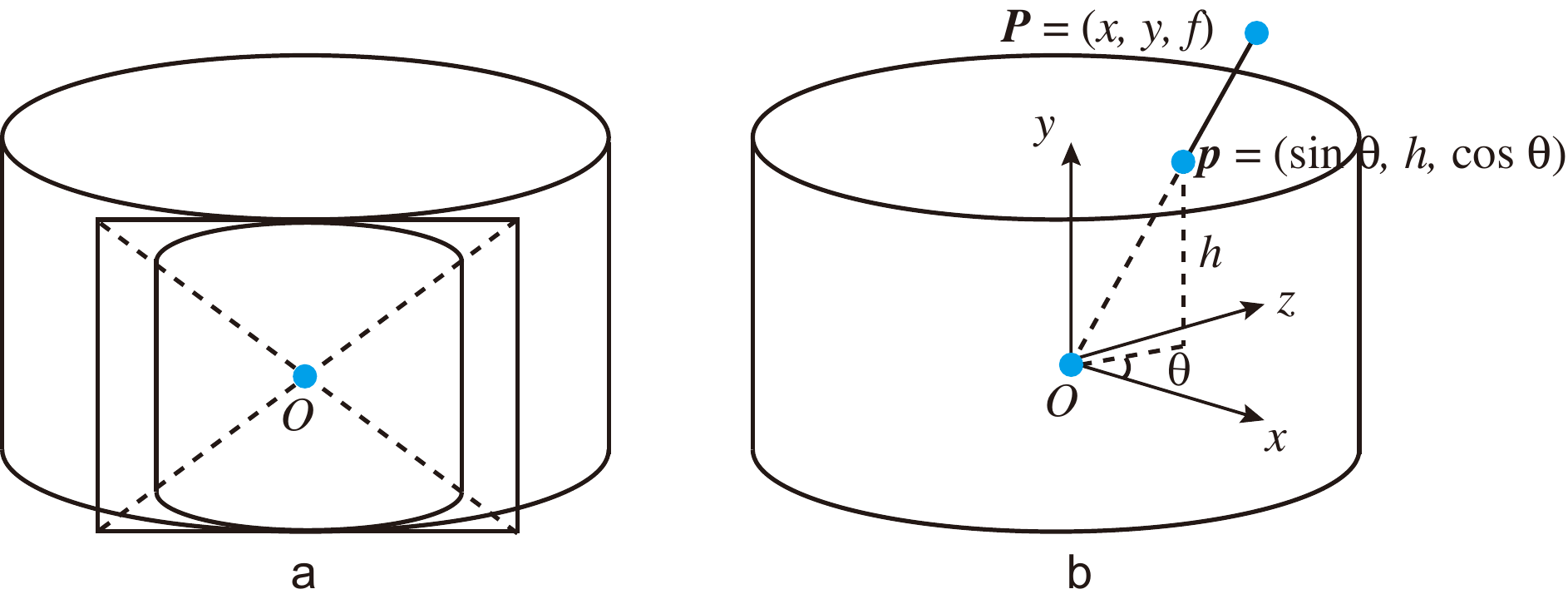}
	\caption{Illustration of cylindrical warp. (a) an image before and after projecting onto the cylinder by center $\bm{O}$. (b) a point $\bm{P}$ from real world to $\bm{p}$ in cylinder \cite{5}.}
	\label{cyl}
\end{figure}

Cylindrical warp \cite{4} is often applied to image stitching to reduce the projective distortion produced by homography and get a multi-view panorama when the camera is assumed to be rotating around its vertical axis. A sketch of cylindrical warp is illustrated in Fig. \ref{cyl}. Generally, let $f$ be the focal length of the cylinder, then a plane 3D point $(x,y,f)$ is corresponding to the 3D cylindrical points $(\sin\theta,h,\cos\theta)$, and the 2D point $(x',y')$ after cylindrical warp $\mathcal{C}$
is computed by
\begin{equation}
x'=f\tan^{-1}\frac{x}{f},~y'=f\frac{y}{\sqrt{x^2+f^2}}.
\end{equation}
Suppose $(a_0,b_0)$ to be the center of the cylinder, then
the warping function $C(x,y)$ can be rewritten as
\begin{align}
x'&=C_x(x,y)=f\tan^{-1}\frac{x-a_0}{f}+a_0, \label{cx}\\
y'&=C_y(x,y)=f\frac{y-b_0}{\sqrt{(x-a_0)^2+f^2}}+b_0. \label{cy}
\end{align}

Note that if $x=a_0$ is a fixed constant, then both $C_x(x,y)$ and $C_y(x,y)$ are linear functions of $y$,
\begin{equation}
C_x(a_0,y)=x_0,~C_y(a_0,y)=y.
\end{equation}
Cylindrical warp maps the vertical line $l_y=\{(a_0,y)|y\in\mathbb{R}^2\}$ to $\{(a_0,y)|y\in\mathbb{R}^2\}$ (itself), which preserves the value of points and ratio of lengths on this line. In this manner, we can divide the image plane into two regions by the line $l_y$, and separately manipulate homography warp and a composition of homography and cylindrical warps on each region.

\begin{figure}[h]
	\centering
	\includegraphics[width=0.48\textwidth]{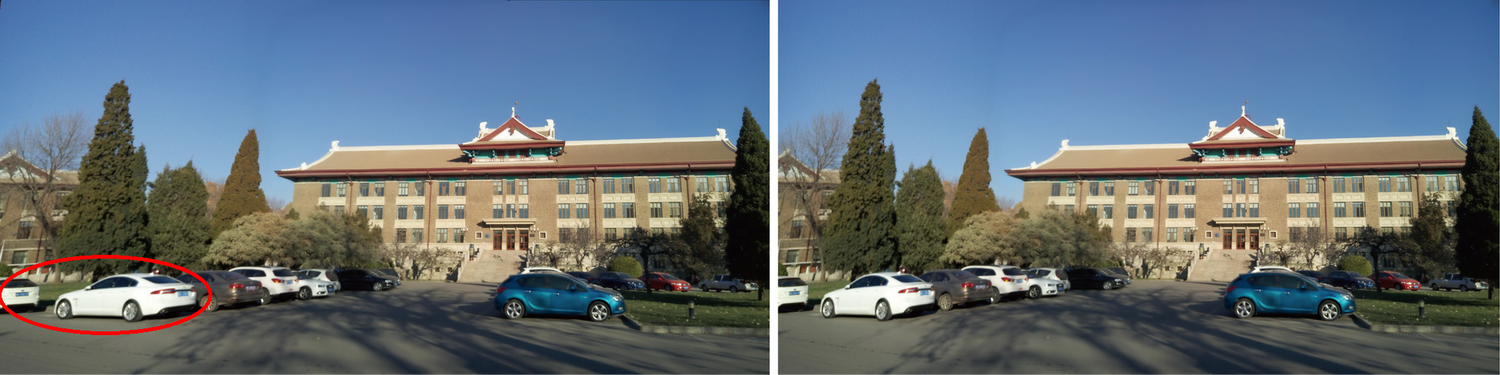}
	\caption{before (left) and after (right) pixel selection in horizontal}
	\label{two}
\end{figure}

Given an image $I$ with height $h$ and width $w$, we let the upper-left point to be the origin, and $x$-axis towards right, $y$-axis towards down. The warping function of half-cylindrical warp $\mathcal{T}$ is defined as
\begin{equation}
T(x,y)=
\begin{cases}
H(x,y),\ \text{if}\ (x,y)\in R_H,\\
C\circ H(x,y),\ \text{if}\ (x,y)\in R_C,
\end{cases}
\end{equation}
where $H(x,y)$ denotes the homography warp, $R_H$ and $R_C$ are two regions divided by a line to apply different warps. The composition of homography and cylindrical warps efficiently mitigates the projective distortion in vertical, however, there is still some horizontal stretches or squeezes which makes the image unnatural, see Fig. \ref{two}.

\section{Pixel selection for ratio-preserving}
The composition of homgraphy and cylindrical warps can only reduce the vertical distortion efficiently. A novel pixel selection strategy is proposed to preserve image width when employing warping function $\mathcal{T}$ in the non-overlapping region. Without ambiguity, we use $(x,y)$ to denote original points, $(x',y')$ to denote points after homography, $(x'',y'')$ to denote points after composition of homography and cylindrical warps. Our strategy is proceeded by:
\begin{enumerate}
	\item Using feature points to estimate a similarity warp
	\begin{equation}
	S = \left(\begin{array}{ccc}
	a & -b & t_x\\
	b& a & t_y
	\end{array}
	\right),
	\end{equation}
	and compute the scale $s$ of pixel selection by $s=\sqrt{a^2+b^2}$, where $t_x$ and $t_y$ are the translation in horizontal and vertical respectively.
	\item For any integer $t\in[1,h]$, selecting the points $\{(x_i,y_i\}_{i=1}^N$ (where $y_i=t$) on horizontal lines among the image domain as sample points, where $N$ and $x_i$ is defined by
	\begin{equation}
	N= \lfloor sw\rfloor,~x_i=\frac{iw}{N}.
	\end{equation}
	Recording these sampling points which are in the non-overlapping region after homography and cylindrical warps by $\{(x''_i,y''_i)\}_{i=1}^N$.
	
	\item Constraining the horizontal distance of these neighboring sample points to be one pixel in the final canvas, then applying the inversion map of cylindrical and homography warps on points $\{(x''_i,y''_i)\}_{i=1}^N$  and manipulating bilinear interpolation when the transformed coordinate is not an integer.
\end{enumerate}
As we can see from the above procedures, the pixel selection strategy preserves the ratio as similarity in the non-overlapping region after $\mathcal{T}$. Moreover, the inversion of warps and interpolation make the results without blurring. Therefore, by combining warp $\mathcal{T}$ and pixel selection strategy we finally get a more natural stitched result with less vertical and horizontal distortion.

\section{Parameter estimation}

It is worth noting that there are some crucial parameters in Eq. (\ref{cx},\ref{cy}), which should be carefully dealt with before applying half-cylindrical warp $\mathcal{T}$. In the following, we only consider horizontal stitching, vertical stitching can be obtained by a similar manner.

Firstly, we need to determine the abscissa $a_0$ of center. Since our final stitched result is divided by a line for applying different warps by its side,
we set $a_0$ to be $1$ or $w$ (depending on the position of the target image). This line also divides the image into overlapping region and non-overlapping region, and the coordinate of homography and cylindrical warp on this line equals each other, thus we ensure the continuity of the stitched image.
\begin{figure}[h]
	\centering
	\includegraphics[width=0.37\textwidth]{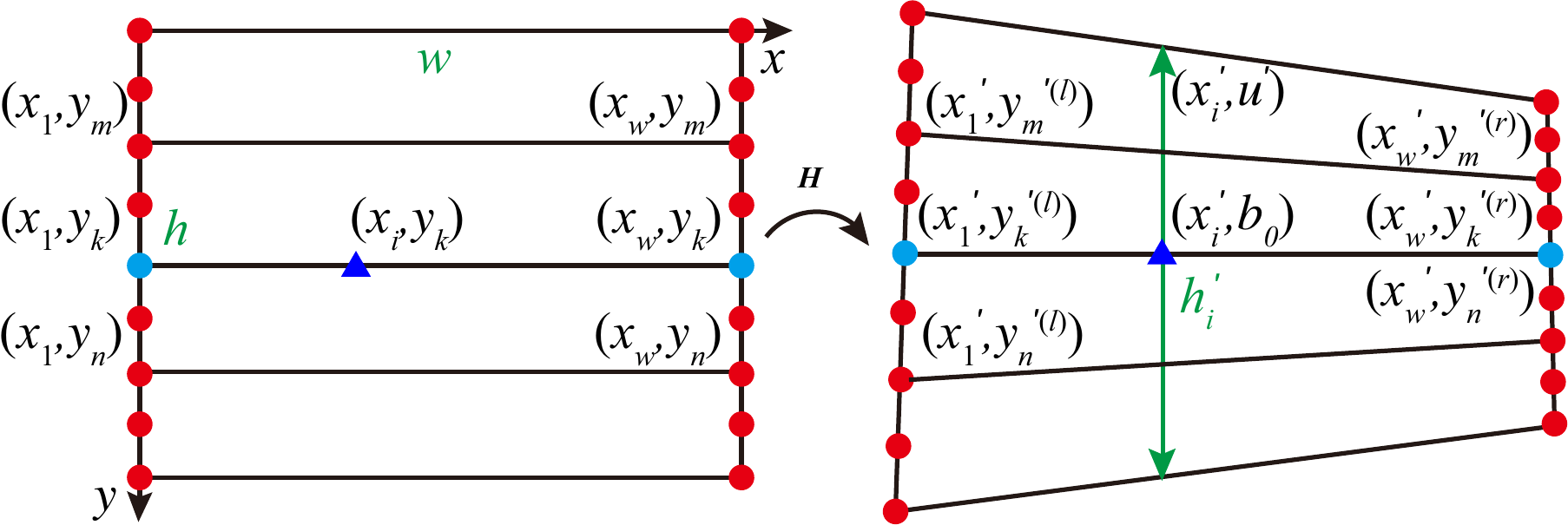}
	\caption{Illustration of selecting $b_0$ with minimal $y$-offset}
	\label{pipeline}
\end{figure}

Then, the ordinate $b_0$ of center should make the pair of corresponding points (e.g., $(x_1,y_m)$ and $(x_w,y_m)$) between vertical boundaries with minimal $y$-offset, as shown in Fig. \ref{pipeline}. We subsequently sample and track the points on the left and right boundaries of original image. Suppose $\{y_{i}'^{(l)}\}_{i=1}^h$ and $\{y_{i}'^{(r)}\}_{i=1}^h$ to be the ordinate of warped points under homography on the left and the right boundary respectively, then $b_0$ is estimated by
\begin{equation}
\hat{b_0}=\frac{y_k'^{(l)}+y_k'^{(r)}}{2},\text{ where }
k=\mathop{\arg\min}_{i}|y_{i}'^{(l)}-y_i'^{(r)}|.
\end{equation}
In fact, there exists a unique horizontal line that remains horizontal, i.e., the $y$-offset equals $0$,  more details can be found in \cite{3}. Our estimation can be regarded as the discretization process of that method.

Finally, an appropriate focal length $f$ should be carefully estimated, since with too large or too small focal length can greatly affect the quality of the final stitched results. The primary purpose of our cylindrical warp is trying to make the image preserve the average height as original as possible after warping. For point $(x'_i,y'_i)$ after homography, the height in position $x'_i$ after the composition of homgoraphy and cylindrical warps, denoted by $h_i''$,  can be derived from Eq. (\ref{cy}) 
\begin{equation}
h''_i=C_y(x'_i,u'+h'_i-1)-C_y(x'_i,u')+1=f\frac{h_i'-1}{\sqrt{(x'_i-a_0)^2+f^2}}+1,
\end{equation}
where $h'_i$ is the height after homography warp in $x'_i$ (see the green  line in Fig. \ref{pipeline}). Then, we estimate focal length $f$ by solving the following non-linear  error minimisation task
\begin{equation}
\hat{f}=\mathop{\arg\min}_{f}\sum_{i=1}^{w}\|h''_i-h_D\|^2,
\end{equation}
where $h_D$ is the height of desired average reference height after cylindrical warp. We set
\begin{equation}
h_D=\max(h,\frac{h_1'+h_w'+2h}{4})
\end{equation}
to preserve the average height after homography. The selection of $h_D$ determines the focal length then further changes the shape of final results.

\section{Experiments}

In our experiments, we employ SIFT \cite{6} to extract and match features, RANSAC \cite{7} to estimate a global homography, and seam-cutting to blend the overlapping region.

\begin{figure}[h]
	\centering
	\includegraphics[width=0.48\textwidth]{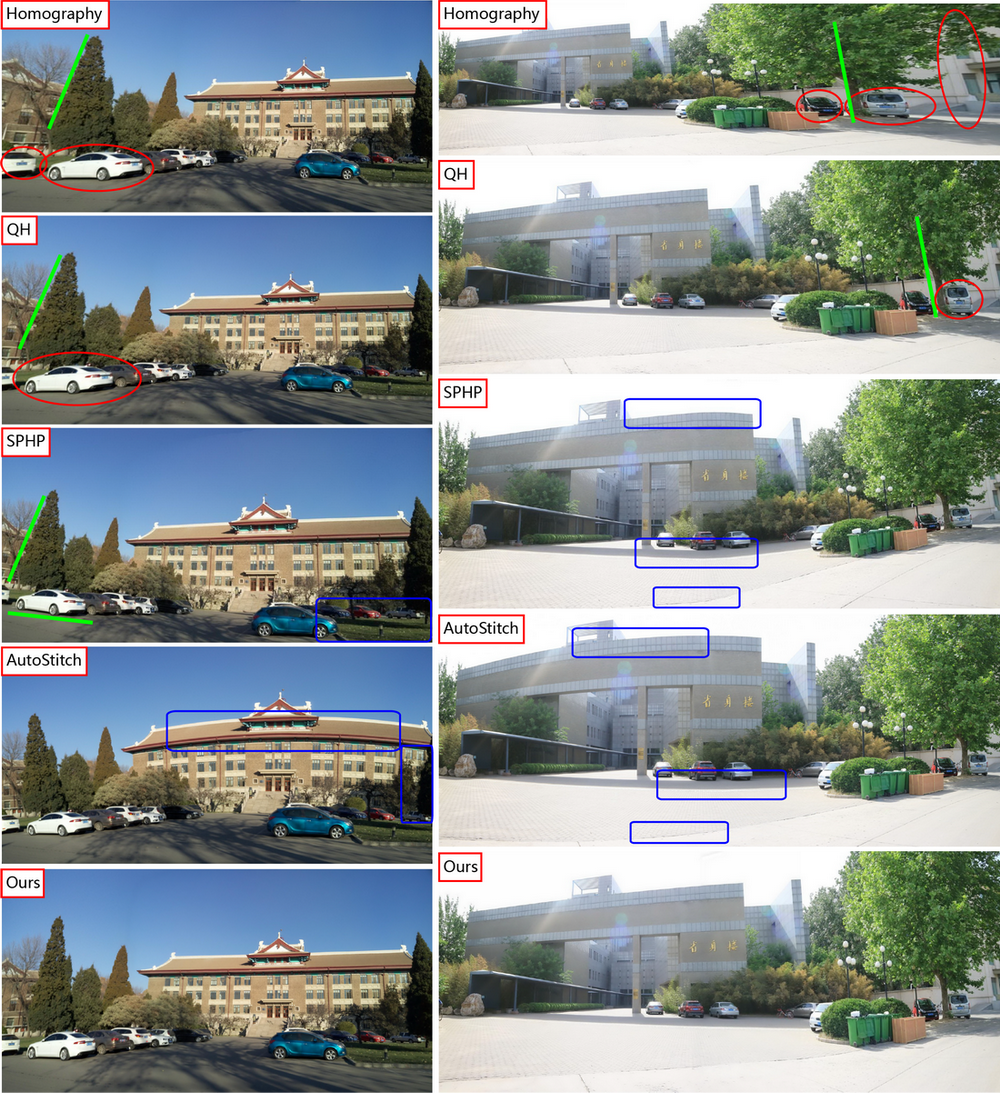}
	\caption{Stitched result comparisons. We use blue rectangles to highlight curved lines and missing content, red ellipses to highlight horizontal distortion, and green lines to highlight vertical distortion}
	\label{result}
\end{figure}

For naturalness, we compare our warp with homography, SPHP \cite{2} and QH \cite{3}. Fig. \ref{result} illustrates the naturalness comparison of stitched results, it is clear that our half-cylindrical ratio-preserving warp looks more natural than other warps in aspects of distortion in horizontal and vertical.

For efficiency, we resize the input images to $1/8$ of original size to speed up the time-consuming seam-cutting stage, then map the seam to original image for blending. Generally, our warp needs very little time, and the average total processing time for different resolutions of proposed method is significantly reduced as shown in Table \ref{time}.
\begin{table}
	\centering
	\caption{Comparisons of average processing time}
	\scalebox{0.9}{{\begin{tabular}{|c|c|c|c|}
			\hline
			resolution (pixel)&  warp time  (s)& \multicolumn{2}{c|}{total time (s)} \\
			\cline{3-4}
			& & original & after resizing\\
			\hline
			1500$\times$2000 &0.09 & 59.71  & 1.43  \\
			\hline
			1920$\times$2560 &0.16 & 127.13 & 2.18  \\
			\hline
			2448$\times$3264 &0.26 & 195.47 & 3.65 \\
			\hline
	\end{tabular}}}{\label{time}}
\end{table}

\section{Conclusions}

In this letter, a novel half-cylindrical ratio-preserving warp utilizing the property of cylindrical warp and a horizontal pixel selection strategy has been proposed to mitigate the projective distortion produced by homography. The parameter estimation stage further guarantees the robustness and naturalness of our warp. Experiments demonstrated that our warp can effectively mitigate projective distortion both in horizontal and vertical, and possesses high time efficiency.

\vskip6pt

%\includegraphics[width=2.5in]{ex_1.jpg}
%\includegraphics[width=2.5in]{ex_2.jpg}

%\bibliographystyle{abbrv}
%\bibliography{plane_cyl}

%\ifCLASSOPTIONcaptionsoff
%  \newpage
%\fi

\bibliographystyle{IEEEtran}
%\balance
%\bibliography{quasi-homography}

\end{document}